\documentclass[10pt, a4paper]{article}
\usepackage{lrec2016}
\usepackage{multibib}
\newcites{languageresource}{Language Resources}
\usepackage{graphicx}
\usepackage[utf8x]{inputenc}

\usepackage{epstopdf}
\usepackage{lingmacros}

\title{Modeling Language Change in Historical Corpora: The Case of Portuguese}

\name{Marcos Zampieri\textsuperscript{1,2}
\hspace{0.9cm}
Shervin Malmasi\textsuperscript{3}
\hspace{0.9cm}
Mark Dras\textsuperscript{3}}

\address{Saarland University, Germany\textsuperscript{1}, German Research Center for Artificial Intelligence (DFKI), Germany\textsuperscript{2}\\
Dept of Computing, Macquarie University, Sydney, NSW,  Australia\textsuperscript{3}\\
\tt marcos.zampieri@dfki.de, shervin.malmasi@mq.edu.au, mark.dras@mq.edu.au\\}

\abstract{This paper presents a number of experiments to model changes in a historical Portuguese corpus composed of literary texts for the purpose of temporal text classification. Algorithms were trained to classify texts with respect to their publication date taking into account lexical variation represented as word n-grams, and morphosyntactic variation represented by part-of-speech (POS) distribution. We report results of 99.8\% accuracy using word unigram features with a Support Vector Machines classifier to predict the publication date of documents in time intervals of both one century and half a century. A feature analysis is performed to investigate the most informative features for this task and how they are linked to language change.\\ \newline }

\begin{document}

\maketitleabstract

\section{Introduction}

It is well-known that language changes over time both in spoken and in written forms. Changes in written language can be manifested in many ways such as the use of the lexicon, grammatical structures, and textual stylistics. Recent studies have shown that it is possible to use language change to predict the approximate publication date of texts in diachronic text collections \cite{ciobanu13,popescu2015semeval}. This task is called temporal text classification and to our knowledge it has not been substantially explored in the literature as other text classification tasks. This paper contributes in this direction. 

In this paper we investigate the use of supervised machine learning classifiers to predict when a text was written using lexical and morphosyntactic information. The classifiers were trained and tested on a sample of a historical Portuguese corpus called Colonia \cite{zampieriandbecker13} which contains texts spanning from the 16\textsuperscript{th} to the early 20\textsuperscript{th} century. The approach we propose here is language independent and it can be applied to any diachronic corpus provided that it is annotated with POS information. 

This study is of interest not only to scholars working in text classification and NLP but also to linguists of different branches, particularly those interested in historical linguistics, and scholars in the digital humanities who often deal with historical manuscripts whose publication date is unknown or uncertain.

This paper is organized as follows: in Section 2 we present related studies that take temporal information and language change in text collections into account (not limited to text classification). In Section 3 we describe the data, the features and the computational approach we used in our experiments; in Section 4 we present the results of three sets of experiments included in this paper. In section 5 we analyze the most informative features in the classification experiments and present a linguistic analysis of important features that indicate language change in the corpus. Finally, Section 6 presents some conclusions and avenues for future work, most notably the question of representing time intervals for temporal text classification.

\section{Related Work}
\label{sec:related}

Modeling temporal information in text is a relevant task to a number of NLP applications. Information Retrieval (IR) methods, for example, often have to process temporal information in both queries and documents to deal with dynamicity of the content found in data repositories and the Web \cite{dakka12,preotiuc2014,kanhabua2015temporal,zhao2015temporal,zhao2015sub}. Time expressions (e.g. \textit{after 2010}), can help algorithms to identify the approximate publication date of texts \cite{chambers2012}, but there are a number of cases in which they are not present in text and one alternative is to use features related to language change as we propose in this paper. 

As will be evidenced in this section, even though there were a number of attempts to approach temporal text classification, to our knowledge this task was not substantially explored as other text classification tasks. The work by \newcite{dejong05} uses unigram language models combined with smoothing techniques and log-likelihood ratio measure (NLLR) \cite{kraaij04} to classify documents within different time spans. The method was tested on a collection of Dutch journalistic texts published from January 1999 to February 2005. Other methods, such as \newcite{kumar11}, make use of information gain to estimate the best features in classification. In Dalli and Wilks \shortcite{dalli06} researchers train a classifier to predict the publication date of texts within a time span of nine years. The method uses words as features and it is aided by words which increase their frequency at some point of time, particularly named entities. Another study that works under a similar assumption is the one published by \newcite{abe10}. The authors proposed the use of similarity metrics to categorize texts based on keywords calculated using tf-idf (term frequency - inverse document frequency). 


\newcite{garcia2011} presents a method to predict the publication dates of excerpts of French journalistic texts containing between 300 and 500 tokens, published between 1801 and 1944. The corpus used was provided by the organizers of the DEFT2011 challenge \cite{grouin11} which was essentially a temporal text classification task for French following a similar DEFT2010 challenge that included both diachronic and diatopic (regional) variation \cite{grouin10}. \newcite{garcia2011} report 14\% accuracy in predicting the year of publication of texts and 42\% accuracy in predicting the correct decade of publication.

Lexical changes are regarded to be an important feature of diachronic text collections and researchers have proposed methods to track meaning change over time \cite{frermann2016}. The study by \newcite{mihalcea12} investigates how word meanings change over time in three major periods in time: 1800, 1900 and 2000. \newcite{popescu2013} look at significant changes in the use of words across time for the purpose of characterizing epochs using the Google N-Gram collection from 1614 to 2009. \newcite{ciobanu13} and \newcite{ciobanu13b} applied SVM and Random Forest algorithms to classify texts of a historical Romanian text collection regarding their publication date. The authors concluded that the use of lexical features is the best source of information for this task. 

An important issue to take into account when working on temporal text classification is how to represent time. Most studies, including our own, model the task as supervised classification in which algorithms are trained to assign texts to an $n$ number of classes. Each of these $n$ classes represent an arbitrarily defined time interval, for example: a month, a year, or a decade. However, there have been a few attempts to approach this task without relying on predefined time spans. The study by \newcite{niculae14} approached the task using ranking and pairwise comparisons to predict for each pair of documents which one is older and finally to produce a rank of all documents in a collection from older to newer. Another recent study to tackle the issue of time intervals is \newcite{efremova2015}. In this study authors apply clustering methods to automatically obtain optimal time partitions in a dataset of historical Dutch notary acts. We return to this question in Section \ref{sec:timespan} of this paper. 

The style of texts also changes over time and it can be a good indicator to predict the publication date of a document. In \newcite{stajnerandzampieri13} researchers used the style of texts calculated using readability scores to predict the publication date of Portuguese texts in the Colonia corpus. Another related study is the one by \newcite{hughes2012} that investigates the evolution of the style of 537 authors of the Project Gutenberg collection by looking at the usage of grammatical words.


The most recent initiative on temporal text classification is the Semeval 2015 Task 7 `Diachronic Text Evaluation' (DTE).\footnote{Results and methods are described in detail in the shared task report \cite{popescu2015semeval}.} In this shared task the organizers proposed three sub-tasks, two of them consisted of temporal text classification, and a third one dealing with the recognition of time specific phrases. For this task, the organizers compiled and released a test set containing English journalistic texts from 1700 to 2010. Texts were labeled with their approximate publication date in coarse, medium and fine-grained intervals representing six, twelve and twenty years respectively. The task proved to be a very challenging one and the only team to participate in all three sub-tasks was the IXA team \cite{salaberri2015} who used external resources such as Google N-grams and Wikipedia Entity Linking to accomplish the task.  The best performing system in the DTE task was the UCD team \cite{szymanski2015ucd} who achieved 54.2\% precision in identifying the publication date of texts in an interval of 20 years (sub-task 2) using Support Vector Machine (SVM).

\section{Methods}
\label{sec:methods}

Following the results obtained by supervised learning approaches at the SemEval DTE task, in this paper we approach the task using supervised single-label multi-class classification. To test our method we used a Portuguese historical corpus, the aforementioned Colonia\footnote{http://corporavm.uni-koeln.de/colonia/index.html}, and we attribute to each text in the corpus a label corresponding to the time interval in which the text was written. As features we use the lexicon arranged as bag-of-words or word n-grams, and morphosyntatic information represented by POS tags. 

First we present a preliminary experiment using a small sample of the data containing excerpts of around 2,000 tokens each. The same sample was previously used in another temporal text classification approach relying on stylistic and readability features \cite{stajnerandzampieri13}. In this experiment we compared the performance of two machine learning classifiers, Multinomial Naive Bayes (MNB) \cite{frank06} and SVM \cite{joachims2006}.


Secondly we present the main experiments of this paper using a linear SVM classifier and a larger sample of the corpus. In particular, we use the LIBLINEAR\footnote{http://www.csie.ntu.edu.tw/\%7Ecjlin/liblinear/}
package \cite{LIBLINEAR} which has been shown to be efficient for large-scale text classification problems such as this \cite{malmasi:2015:mnli}. As to the sample we used in this experiment, we opted to generate artificial documents composed of mixed sentences from different texts of the same period. 
%

\subsection{Data and Features}
\label{sec:data}

The Colonia corpus is a historical Portuguese corpus, which contains texts spanning from the 16\textsuperscript{th} century to the early 20\textsuperscript{th} century \cite{zampieriandbecker13}. The corpus contains 100 documents (full novels or text collections) amounting to over 5.1 million tokens. It contains sentence boundary mark-up and coarse-grained POS annotation carried out using TreeTagger \cite{schmid94}.\footnote{The corpus description paper does not contain any evaluation regarding the performance of the tagger on the Colonia dataset. The authors addressed solely the question of unknown lemmas after annotation in a post-processing stage.} Colonia is available for download and it can be accessed through different corpus processing tools such as CQPWeb (through the project's website), Linguateca,\footnote{http://www.linguateca.pt/acesso/corpus.php?corpus=COLONIA} and Corpuseye.\footnote{http://corp.hum.sdu.dk/cqp.pt.html} To our knowledge, the corpus has been used to study different aspects of the evolution of Portuguese such as diachronic morphology \cite{nevins2015}.

The availability of suitable texts is a known shortcoming in the compilation of historical corpora. Although Colonia is to our knowledge the biggest Portuguese corpus of its kind, it does not contain many texts from each period. The number of documents varies between 13 from the 16\textsuperscript{th} century and 38 from the 19\textsuperscript{th} century. For text classification, however, the number of documents available (especially at the training stage) is very important to provide enough information to achieve high classification performance.

To circumvent this limitation, in this paper we propose the use of composite documents made of sentences from various texts. Following the methodology of
\newcite{malmasi:2014:cnli}, we randomly select and combine the sentences from the same class (time period) to generate artificial texts of approximately $330$ tokens on average, creating a set of documents for training and testing. This methodology ensures that the texts for each class are a mix of different authorship styles and topics. It also means that all documents are similar and comparable in length making the task more challenging. Previous work in other text classification tasks has shown that longer texts can be easier to classify \cite{malmasi-et-al:2015:adi}.

In our experiments we model two dimensions of language variation across time, lexical and (morpho-)syntactical. We do so by extracting words and part-of-speech (POS) tags from the corpus and using them as features. To the best of our knowledge these features were not yet tested in multi-class temporal text classification for Portuguese. The most similar approach to use these features is the ranking approach proposed by \newcite{niculae14}.

\section{Results}
\label{sec:Results}

In this section we present the results obtained in three sets of experiments: 

\begin{enumerate}
\item In Section 4.1 we describe preliminary experiments using a small sample of the corpus containing 87 documents spanning from the 17\textsuperscript{th} to the early 20\textsuperscript{th} century. We train two algorithms (SVM and MNB) using both words and POS tags represented as bag-of-words to predict the century in which the text was published.
\item In Section 4.2 we apply the method of 
\newcite{malmasi:2014:cnli} to generate artificial composite documents for training and testing using the complete set of texts available in Colonia  (from the 16\textsuperscript{th} to the early 20\textsuperscript{th} century). We train an SVM classifier to predict the century in which the text was published and given the substantial increase in training material we report an important increase in performance using POS tags or words represented as uni-, bi-, and trigrams.
\item Finally, in Section 4.3 we replicate the methods used in Section 4.2 for a smaller time span of 50 years.
\end{enumerate}

\subsection{Preliminary Experiments}
\label{sec:Preliminary}

As the preliminary experiments of this paper we use a bag-of-words model on a sample of the data previously used in the aforementioned study by \newcite{stajnerandzampieri13}. Each document in this sample contains up to 2,000 token. The criteria for sampling was inspired in the Brown family corpora \cite{francis1979brown}. The distribution of the texts across centuries is presented next.

\vspace{2mm}

\begin{table} [h!]
\begin{center}
\label{tab:Corpora}
\scalebox{0.95}{
\begin{tabular}{c c c}

\hline
{\bf Century} & {\bf Texts} & {\bf Tokens} \\
\hline
17th & 18 & 31,635\\
18th & 14 & 23,175\\
19th & 38 & 63,950\\
20th & 17 & 28,569\\ \hline
Total & 87 & 147,329\\
\hline
     
\end{tabular}
}
\caption{Colonia Sample: Preliminary Experiments Using Text Excerpts of Less Than 2,000 Tokens}
\end{center}
\end{table}

\vspace{-2mm}


     


In Colonia there are not many texts from each class and only a few from the 16\textsuperscript{th} century. For this reason, in \newcite{stajnerandzampieri13} and in this preliminary experiment, we disregard texts from this century and propose an experiment with four classes instead of the five represented in Colonia. In Table \ref{resultspreliminary} we present accuracy results using $k$-fold cross-validation, with $k = 10$. We considered the majority class (19\textsuperscript{th} century) as the baseline performance.

\vspace{2mm}

\begin{table}[h]
\centering
\scalebox{0.95}{
\begin{tabular}{l l r}
      \hline
      {\bf Algorithm} & {\bf Features} & {\bf Accuracy (\%)} \\ \hline
	Baseline & & $43.5$\\ 
    \hline
    MNB &	Words + POS		& $72.5$ \\ 
	MNB &	Words 			& $70.5$ \\ 
	MNB & 	POS				& $66.1$  \\
	 \hline
	SVM &	Words + POS		& $74.1$ \\ 
	SVM &	Words 			& $72.5$ \\ 
	SVM &	POS				& $67.4$ \\
	 \hline

\end{tabular}
}
\caption{Preliminary Experiments: Results}
\label{resultspreliminary}
\end{table}


The best results were obtained by SVM regardless of the features used. SVM achieved 74.1\% accuracy in identifying the century of texts using a combination of words and POS tags. An expected outcome of these preliminary experiments is that the results using lexical and morphsyntactic features are substantially higher than the 59\% accuracy reported by \newcite{stajnerandzampieri13} using readability/stylistic features.



\subsection{Increasing the Sample}
\label{sec:sample}

The small number of texts is a known limitation of most historical corpora. We address this question by generating artificial data with the methods described in Section 3. 

In this experiment we used a set of 1,500 artificially generated documents each of them combining sentences of various texts to represent each class.\footnote{This was the largest possible number that would yield an even distribution across classes.} Details of the final data used in this experiment are shown in Table \ref{tab:gendata}.

\begin{table}[!ht]
\centering
\scalebox{0.95}{
\begin{tabular}{c c c c}
\hline
{\bf Century} & {\bf Texts} & {\bf Tokens} \\
\hline
16th	& 1,500 & 507,848\\
17th	& 1,500 & 507,970\\
18th	& 1,500 & 501,506\\
19th	& 1,500 & 483,150\\
20th	& 1,500 & 480,575\\
\hline
Total	& 7,500 & 2,481,049\\
\hline
\end{tabular}
}
\caption{Colonia Sample: Artificial Documents Generated Divided by Century}
\label{tab:gendata}
\end{table}

Results obtained using an SVM classifier are presented in Table \ref{tab:results}. Word $n$-grams of order $1$--$3$ and POS $n$-grams of order $1$--$3$ were extracted and a single SVM classifier was trained on each of these six $n$-gram features. Using the same evaluation methods of the preliminary experiment, we report accuracy results under $k$-fold cross-validation, with $k = 10$. Results are compared with a random baseline of 20\%.

\vspace{4mm}

\begin{table}[!ht]
\center
\scalebox{0.95}{
\begin{tabular}{lr}
\hline
\textbf{Feature} & \textbf{Accuracy (\%)} \\
\hline
Baseline & $20.0$\\
\hline
Word unigrams	& $99.8$\\
Word bigrams	& $98.9$\\
Word trigrams	& $96.2$\medskip\\
Part-of-Speech unigrams	& $69.1$\\
Part-of-Speech bigrams	& $87.3$\\
Part-of-Speech trigrams	& $90.7$\\
\hline
\end{tabular}
}
\caption{Increasing the Sample: Results}
\label{tab:results}
\end{table}

The best results were obtained using word unigrams achieving 99.8\% accuracy. Results are substantially higher than those obtained in the preliminary experiment, once more confirming that the amount of training material plays a crucial role in this task. We look in more detail to the most informative features in classification in Section \ref{sec:Feature}. 

\begin{figure}[!ht]
\centering
\includegraphics[width=.52\textwidth]{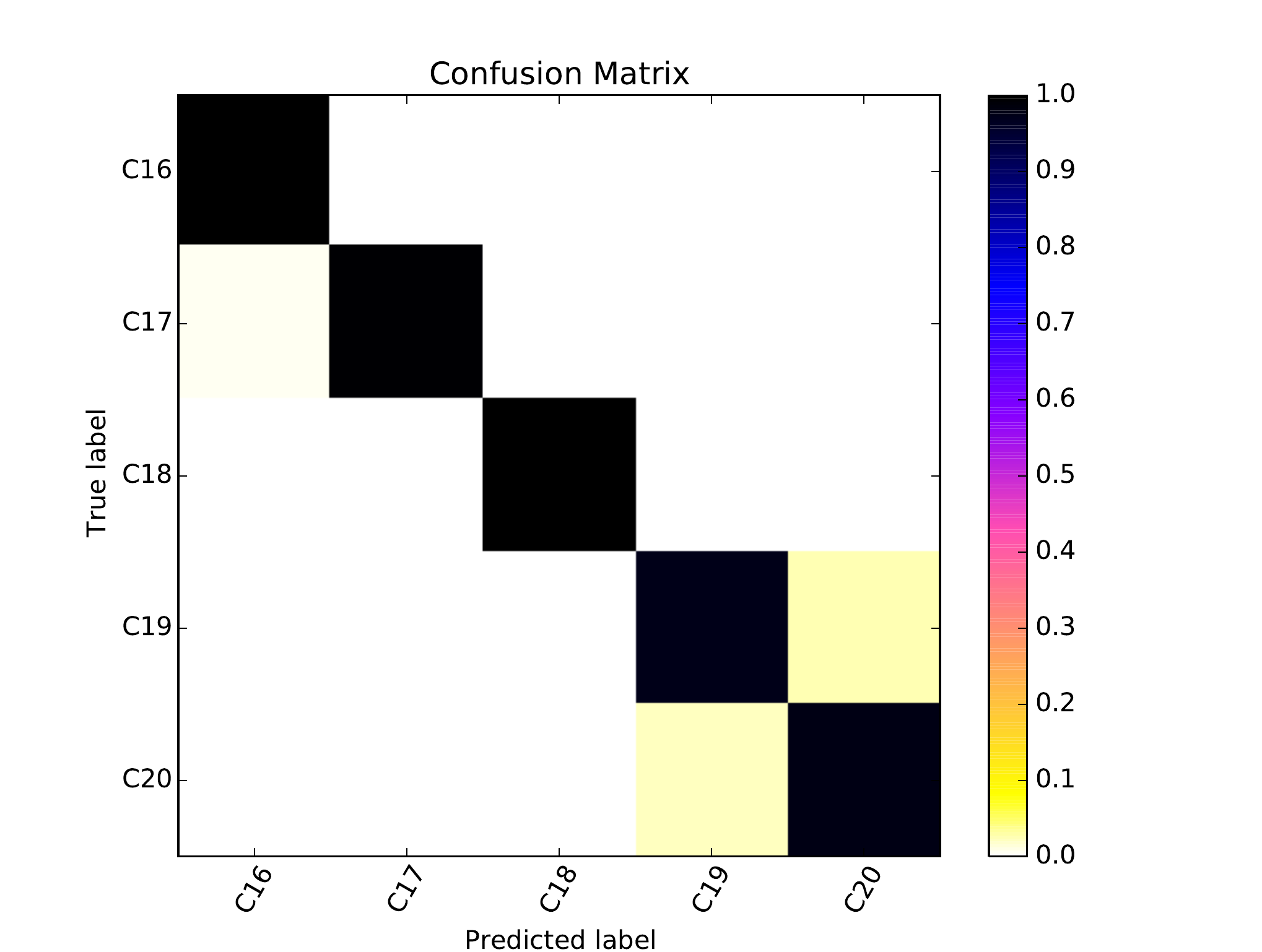}
\caption{Confusion Matrix: Century Classification}
\label{fig:lc}
\end{figure}

Next we present the confusion matrix obtained by the best performing setting, word unigrams, in Figure~\ref{fig:lc}. We can observe that the classification performance is perfect for almost all centuries, with only some small confusion between the 19\textsuperscript{th} century and early 20\textsuperscript{th} century. 

We contend that merging segments from different texts to form artificial documents tends to decrease the importance of stylistic preferences, lexical choices and other idiosyncrasies of a particular author. A result that corroborates this hypothesis is the performance obtained by the classifier using POS tags. The method is able to predict the century of the documents relying on POS trigrams with 90.7\% accuracy. In our opinion, this is an indication that there are important structural properties that differ in each of these time spans. This is an indication that the algorithm is able to capture language variation beyond word forms, as noted by \cite{zampierietal13} in a similar classification experiment to study the diatopic variation of Spanish.

\subsection{Smaller Time Intervals}
\label{sec:small}

As our method obtained almost perfect performance in predicting the century of each text, we would like to evaluate its performance in predicting the publication date of texts using a shorter time interval. For this purpose We divided the documents in the corpus into time intervals of 50 years resulting in 9 classes. This also results in a substantial drop in the random baseline, compared to the previous experiment. We then used the same methodology for generating artificial documents resulting in a total of 450 documents per class,\footnote{Adopting the same criterion used in the previous section, we used the largest possible number of documents that would result in an even distribution across classes.} 4,050 documents in total and 1.35 million tokens. The distribution of data along with the total token count per time interval is presented in Table \ref{tab:gendatasmall}.

\vspace{2mm}

\begin{table} [ht]
\centering
\scalebox{0.95}{
\begin{tabular}{c c c c}
\hline
{\bf Time Interval} & {\bf Texts} & {\bf Tokens} \\
\hline
1500-1550	& 450 & 156,087\\
1551-1600	& 450 & 151,173\\
1601-1650	& 450 & 154,347\\
1651-1700	& 450 & 152,577\\
1701-1750	& 450 & 141,491\\
1751-1800	& 450 & 161,804\\
1801-1850	& 450 & 148,608\\
1850-1900	& 450 & 145,134\\
1901-1950	& 450 & 143,597\\
\hline
Total	& 4,050 & 1,354,818\\
\hline
\end{tabular}
}
\caption{Colonia Sample: Artificial Documents Generated Divided by Time Intervals of 50 Years}
\label{tab:gendatasmall}
\end{table}

We used the sample presented in Table \ref{tab:gendatasmall} for automatic classification and the results are presented in Table \ref{tab:resultssmall}. Once more, our best performing setting achieved 99.8\% accuracy using word unigrams. For the other settings we observed a slight (and expected) decrease in performance, with the lowest result obtained using POS unigrams 2.2 percentage points lower than the century classification.

\begin{table}[!ht]
\center
\scalebox{0.95}{
\begin{tabular}{lr}
\hline
\textbf{Feature} & \textbf{Accuracy (\%)} \\
\hline
Baseline & $11.0$\\
\hline
Word unigrams	& $99.8$\\
Word bigrams	& $98.7$\\
Word trigrams	& $93.8$\medskip\\
Part-of-Speech unigrams	& $66.9$\\
Part-of-Speech bigrams	& $85.7$\\
Part-of-Speech trigrams	& $90.1$\\
\hline
\end{tabular}
}
\caption{Smaller Time Intervals: Results}
\label{tab:resultssmall}
\end{table}

It is interesting to note that even when working with shorter time spans, the distinction based on POS distribution tends to perform quite well which suggests that there are structural differences between centuries as well as between time spans of 50 years. 

\begin{figure}[!ht]
\centering
\includegraphics[width=.48\textwidth]{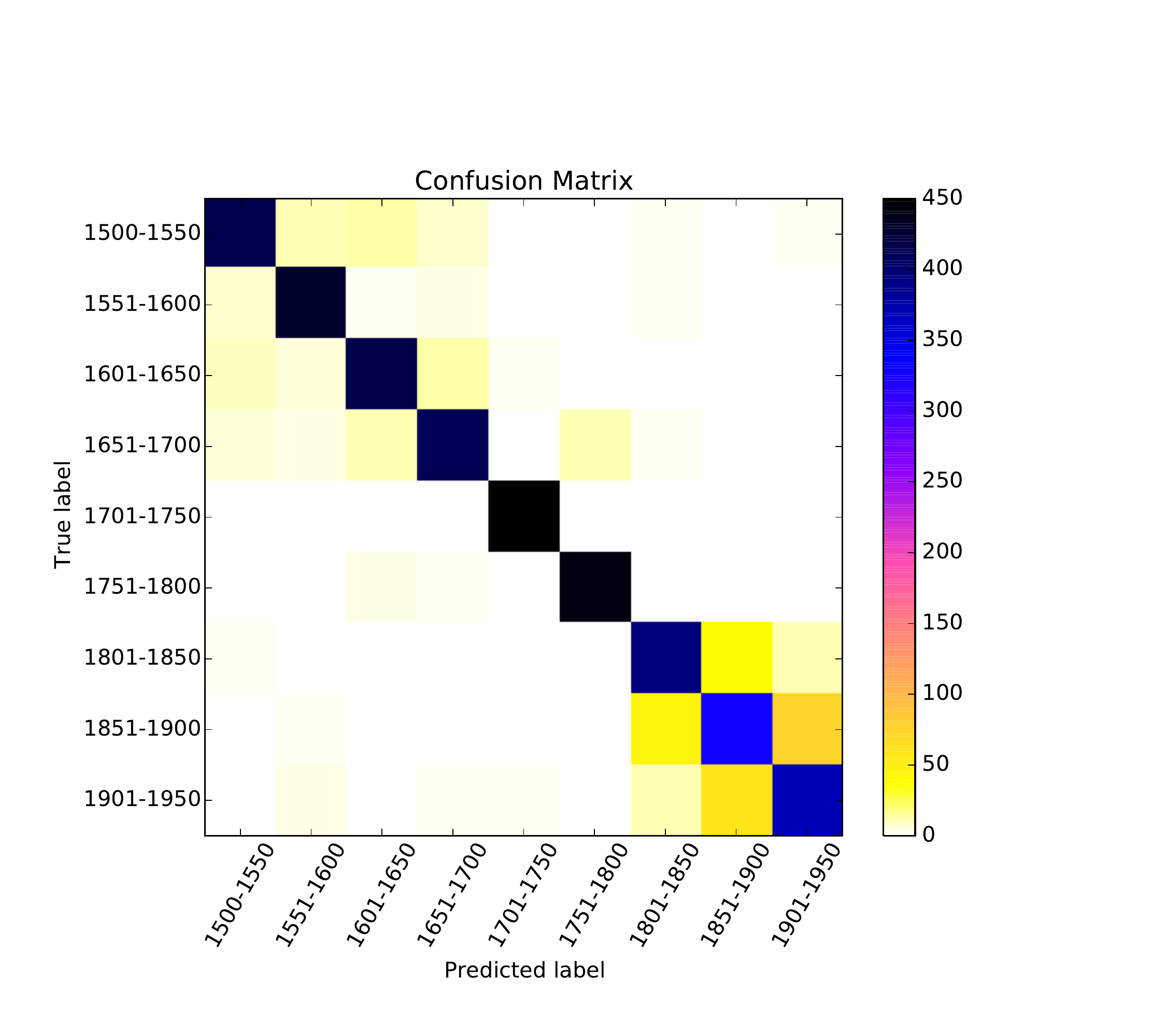}
\vspace{-0.5cm}
\caption{Confusion Matrix: Time Intervals of 50 Years using POS trigram features.}
\label{fig:cm50}
\end{figure}

In figure \ref{fig:cm50} we present the confusion matrix obtained by the best setting using POS tags as features, POS trigrams, which achieved $90.1\%$ accuracy. The results indicate that whenever a set of time intervals poses more challenge to the classifier, the more similar they are in terms of grammatical structures. We observed some degree of confusion in the oldest part of the corpus from 1500 to 1600 and from 1600 to 1700. The most challenging set of time intervals was the period comprising the 19\textsuperscript{th} and 20\textsuperscript{th} century. We take a closer look at indicative features of language change in Section 5.

\vspace{2mm}

Another interesting pattern we observed is that performance variation across different sets of features is almost identical for experiments described in Sections 4.2 and 4.3 here. Word-based methods perform best when arranged as unigrams and decrease performance for bigrams and even more when arranged as trigrams. On the other hand, classification using POS tags obtain higher performance using trigrams than bigrams and unigrams. This seems intuitive, once lexical variation is often captured by looking at individual words whereas structural differences tend to be observed when looking into sequences of two or more words.

\section{Indicative Features of Language Change}
\label{sec:Feature}

In this section we look at the most informative features for each of the five centuries represented in Colonia and try to highlight patterns that are, or are likely to be, good indicators of language change. 

The most informative features were extracted using the methodology proposed in  \newcite{malmasi:2014:lth}. This works by ranking the features according to the weights assigned by the SVM model. In this manner, SVMs have been successfully applied in data mining and knowledge discovery in a wide range of tasks such as identifying discriminant cancer genes \cite{guyon2002gene}.

As noted by \newcite{zampierietal13}, in a similar text classification task involving diatopic variation, linguistically motivated features usually do not outperform word- and character-based features. Our results also corroborate this claim. Even so, the use of features represented by POS tags and/or morphological information may provide interesting insights on language variation that can be analyzed by looking at the output of classifiers. 

We observed that constructions including three verbs, annotated as {\em V V V}, are a highly discriminating feature for 20\textsuperscript{th} century texts. This represents constructions similar to Examples\footnote{All examples were obtained from the Colonia corpus.} 1 and 2, most of them with two auxiliary verbs and a main verb in the past participle tense:


\enumsentence{O major tinha razão: o Leonardo não {\bf parecia ter nascido} para emendas. {\em (EN: seems to be born)} }

\enumsentence{A música como a leitura, {\bf deve ser ministrada} com prudência. {\em (EN: must be administered)}}

Another interesting finding is the overuse of adjectives in the 19\textsuperscript{th} and particularly in the 20\textsuperscript{th} century. As noted in Section 4.3. this is a period in which the classifier has difficulties predicting the publication date of texts. By looking at the most informative trigrams we found patterns such as {\em ADJ NOM ADJ} (adjective noun adjective), and {\em NOM ADJ CONJ} (noun adjective conjunction). We investigated the latter and discovered that the conjunctions in this pattern usually refer to coordinate conjunctions that indicate the use of a second adjective modifying the noun in the same sentence as in Example 3:

\enumsentence{Nada sorria naquela {\bf habitação árida e velha.} {\em (EN: dry and old housing)}}

The {\em ADJ NOM ADJ} (adjective noun adjective) pattern reflects the possibility of using adjectives after or before nouns depending on stylistic preferences or on what the speaker wants to emphasize. In Portuguese it is possible to say both {\em homem grande} and {\em grande homem}. The first one is the literal meaning, {\em big man} or {\em tall man}, whereas the second is a figurative one referring to a man who possesses great qualities, as the English expression {\em great man}. An example of the {\em ADJ NOM ADJ} pattern found in the corpus is presented in 4:

\enumsentence{Os {\bf grandes olhos azuis}, meio cerrados, às vezes se abriam languidamente como para se embeberem de luz, e abaixavam de novo as pálpebras rosadas. {\em (EN: big blue eyes)}}

Due to its thematic relevance, named entities (e.g. location, person) play an important role in text classification. Linguistically, these features are often not a very relevant indication of language change as the other features we discuss in this section, but they are usually very informative for classifiers. In the 17\textsuperscript{th} century names of Portuguese monarchs such as {\em Dom Afonso} and {\em Dom João} are very informative, and the pronoun {\em Sua Majestade (EN: Your Majesty)} is also a very prominent feature.

In the 16\textsuperscript{th} century we observed the use of a number well-document archaisms used in that period \cite{castro91}. This includes most notably the cases of {\em per}, {\em asi}, {\em mui}, {\em mi}, and {\em despois} that refer to the current forms {\em por (EN: for)}, {\em assim (EN: thus, therefore)}, {\em muito (EN: much)}, {\em mim (EN: me)} and {\em depois (EN: after)} respectively. 

The case of {\em per} is particularly interesting because it represents both the lusophone archaism as well as the Latin word from which we can trace its origin. There are a number of quotes in Latin in the 16\textsuperscript{th} century part of the corpus making {\em per} a highly discriminating feature for this century. This can be observed in Examples 5 and 6:

\enumsentence{A segunda, que esse lugar esteja em sítio acomodado pera socorrer dele com facilidade suas conquistas, e fazer as armadas que convém; isto se prova {\bf per} muitas razões.}

\enumsentence{Siquidem me fecistis, constituamos leges, {\bf per} quas terra nostra sit in pace.}

\section{Conclusion}
\label{sec:Conclusion}

We investigated the use of words and POS tags to model lexical and syntactic variation in historical corpora for the purpose of temporal text classification. Our work extends the common knowledge in the task by using lexical and POS information for the first time in multi-label Portuguese temporal text classification, and by using artificially generated test and training instances combining fragments of texts written in the same period. The use of the latter was to the best of our knowledge still not investigated for this task. We contend that this methodology is an interesting strategy to cope with small amount of available texts, which is a known limitation of many historical corpora.

The approach proposed in this paper is able to predict the publication date of texts, in intervals of both 100 and 50 years, using word unigrams with 99.8\% accuracy. The method is also able to predict the publication date of texts using solely POS tags achieving performance of 90.7\% accuracy for intervals of 100 years, and 90.1\% accuracy for intervals of 50 years.  

Finally, we used the most informative features in the classification to investigate indicators of language change.
The high word unigram results could be the result of chronological topic specificity; what's more interesting is the POS n-grams discussed above, which aren't related to topics but are indicative of grammatical or stylistic change.
We found that texts from the 19\textsuperscript{th} and 20\textsuperscript{th} centuries contain on average a larger number of adjectives, which reflect stylistic preferences of that period. This kind of analysis is only possible with the use of POS tags as features. We also showed that, as expected, archaisms are a very distinctive feature of texts from the 16\textsuperscript{th} century.

As future work, we would like to investigate the question of time intervals (see next section) as well as to create an additional test set comprising a few texts from each time period. The new test set can be use to carry out a cross-corpus evaluation, using Colonia to training and test on the new corpus, to provide insights about how the models perform on data from different genres, text types, etc.

\subsection{Future Work: Finding Optimal Time Intervals}
\label{sec:timespan}

To our understanding, one of the limitations of most temporal text classification experiments (including ours) is the arbitrary definition of time intervals. In the case of our dataset, time intervals of one century are too long and they often fail to capture linguistic changes that occur in a certain point in time that do not coincide with the turn of a century. On the other hand, working with too short time spans is often unfeasible for historical datasets as there are not enough data points to be split between a large number of classes. The smallest time interval we could work with using this corpus was 50 years.

Even so, we contend that the definition of arbitrary time spans is valid as a proof of concept and a perfect fit for supervised classification methods that require a predefined set of classes as the SVM-based approach presented in this paper. In real-world tasks, however, one might be interested in using more fine-grained intervals that can better capture the structure of the data and predict the publication date of a document more precisely. In light of this, our efforts are now concentrated on exploring ways to better represent the linearity of time \cite{niculae14} or to find optimal time intervals in historical corpora as proposed by the aforementioned study by \newcite{efremova2015}. We are currently experimenting with a recently proposed clustering method for this purpose \cite{amorim2015}. 

It is important to note, however, that an adaptation of the ranking-based approach by \newcite{niculae14} competed in the SemEval 2015 `Diachronic Text Evaluation' as the AMBRA team \cite{zampieri2015} and did not outperform supervised learning approaches such as the one by the UCD team \cite{szymanski2015ucd}. To our understand this seems to confirm that a combination of supervised classification and other computational approaches to find optimal time intervals (e.g. clustering) is probably the best way to approach this task.

\section*{Acknowledgements}

We would like to thank the anonymous reviewers for providing us important feedback and constructive comments to increase the quality of this paper.

\section*{Bibliographical References}
\label{main:ref}

\bibliographystyle{lrec2016}
\bibliography{temporal}

\end{document}